\documentclass[10pt,twocolumn,letterpaper]{article}

\usepackage[pagenumbers]{cvpr} 

\usepackage{booktabs}
\usepackage{amsmath} 
\usepackage{multirow}
\usepackage{color}
\usepackage{pifont} 
\usepackage{amssymb}
\usepackage{makecell}
\usepackage{float}

\newcommand{\modelname}{PositionIC}
\newcommand{\nerfmodelname}{Visibility-Aware Attention}

\newcommand{\benchname}{PositionIC-Bench}
\newcommand{\datasetname}[1]{PIC-#1}
\newcommand{\datafiltname}{Bidirectional Multi-dimensional Perception Data Synthesis}

\newcommand{\BigName}[1]{\textbf{#1}}

\definecolor{cvprblue}{rgb}{0.21,0.49,0.74}
\usepackage[pagebackref,breaklinks,colorlinks,allcolors=cvprblue]{hyperref}

\def\paperID{6407} 
\def\confName{CVPR}
\def\confYear{2026}

\title{PositionIC: \textbf{U}nified Position and Identity Consistency for Image Customization}

\author{
\BigName{Junjie Hu}\textsuperscript{1,2}\footnotemark[1] \quad
\BigName{Tianyang Han}\textsuperscript{1}\thanks{Equal Contribution.} \quad
\BigName{Kai Ma}\textsuperscript{1}\thanks{Project Leader.} \quad
\BigName{Jialin Gao}\textsuperscript{1} \quad
\BigName{Song Yang}\textsuperscript{1} \quad
\BigName{Xianhua He}\textsuperscript{1} \\
\BigName{Junfeng Luo}\textsuperscript{1} \quad
\BigName{Xiaoming Wei}\textsuperscript{1} \quad
\BigName{Wenqiang Zhang}\textsuperscript{2,3} \quad
\\
\textsuperscript{1}MeiGen AI Team, Meituan \\
\textsuperscript{2}Shanghai Key Lab of Intelligent Information Processing, \\College of Computer Science and Artificial Intelligence, Fudan University, Shanghai, China\\
\textsuperscript{3}College of Intelligent Robotics and Advanced Manufacturing, Fudan University, Shanghai, China\\
}

\begin{document}
\maketitle

\begin{abstract}

Recent subject-driven image customization excels in fidelity, yet fine-grained instance-level spatial control remains an elusive challenge, hindering real-world applications. This limitation stems from two factors: a scarcity of scalable, position-annotated datasets, and the entanglement of identity and layout by global attention mechanisms.
To this end, we introduce \modelname{}, a unified framework for high-fidelity, spatially controllable multi-subject customization. First, we present BMPDS, the first automatic data-synthesis pipeline for position-annotated multi-subject datasets, effectively providing crucial spatial supervision. Second, we design a lightweight, layout-aware diffusion framework that integrates a novel visibility-aware attention mechanism. This mechanism explicitly models spatial relationships via an NeRF-inspired volumetric weight regulation to effectively decouple instance-level spatial embeddings from semantic identity features, enabling precise, occlusion-aware placement of multiple subjects.
Extensive experiments demonstrate \modelname{} achieves state-of-the-art performance on public benchmarks, setting new records for spatial precision and identity consistency. Our work represents a significant step towards truly controllable, high-fidelity image customization in multi-entity scenarios. Code and data: https://github.com/MeiGen-AI/PositionIC.
\end{abstract}

\section{Introduction}
\label{intro}
Diffusion-based models have recently revolutionized visual synthesis, especially in text-to-image (T2I) generation, where they produce photorealistic images closely aligned with short textual prompts~\citep{sdxl,stablediffusion3,ldm,playground,flux}.
Beyond generic generation, subject-driven image customization aims to integrate one or more reference objects into user-specified scenes under the guidance of a text prompt. Traditionally, this task emphasizes two core goals: (i) preserving the appearance identity of the reference object(s), and (ii) aligning the synthesis with the user’s textual description, where remarkable improvements in visual fidelity have been achieved by recent methods~\citep{uno,dreamo,omini_control,wang2025unicombine}.

\begin{figure}[t]
    \centering
    \includegraphics[width=0.9\linewidth]{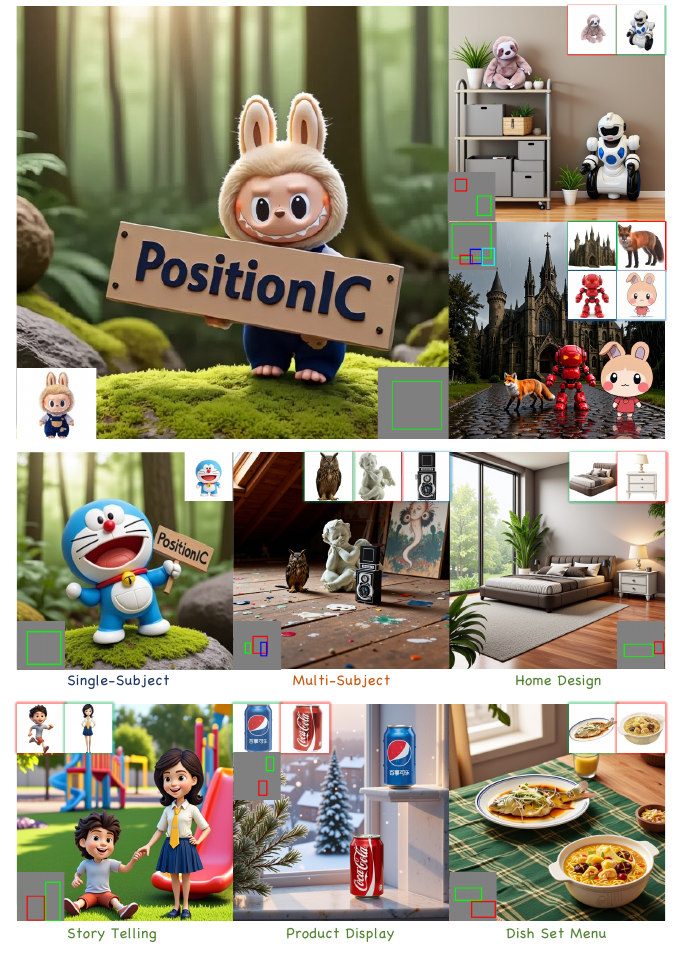}
    \caption{ Results from \modelname{} across various controllable image customization tasks.}
    \label{fig:fig1_1}
\end{figure}
However, through exploring real-world applications like e-commerce product display, storybook illustration, and interior design, we identified an additional critical dimension: fine-grained, flexible spatial control, encompassing aspects such as subject placement, scale, occlusion, and relative layout. In other words, while prior works focus on “\textit{what}” to generate, practical usage demands \textit{where} and \textit{how} each subject appears. To explicitly address this, we adopt a refined task definition: subject-driven customization with both identity fidelity and instance-level spatial controllability. This is a natural extension of the established task, not a fundamentally different objective, but one that better matches application requirements.

Despite substantial progress in identity-preserving personalization methods, current approaches still struggle with flexible placement and layout control. Specifically, two coupled bottlenecks remain: (1) the scarcity of large-scale datasets annotated with explicit positional labels of multiple subjects, hindering the learning of spatial reasoning; and (2) the prevalent use of global-level attention mechanisms in diffusion transformers, which inadvertently entangle semantic identity and spatial layout, thereby failing to provide fine-grained, visibility-aware control. While a few recent works~\citep{ms-diffusion,gligen} have attempted layout control, they typically exhibit a trade-off between spatial precision and subject identity consistency, thus limiting their practical utility.

To this end, we introduce two tightly coupled components: (i) a scalable Bidirectional Multi-dimensional Perception Data Synthesis (BMPDS) pipeline that generates high-quality, multi-subject, position-annotated data; and (ii) \modelname{}, a lightweight layout-aware diffusion framework incorporating a visibility-aware attention mechanism that enables precise, controllable placement of multiple subjects while maintaining identity fidelity.

Specifically, to overcome the data bottleneck, we devise an automatic pipeline that expands single-subject collections into scalable, high-quality multi-subject datasets annotated with explicit position masks. A hierarchical training schedule establishes a bidirectional generation paradigm, progressively moving from single-to-multi subject synthesis and back, so that resolution constraints are relaxed while subject drift is suppressed. Because of the inherent hallucination of Multi-modal Large Language Models (MLLMs)~\citep{han2024instinctive,bpo}, we avoid direct visual comparisons: expert vision models first translate visual content into textual descriptions, after which MLLMs perform multi-dimensional consistency checks. This two-stage filtering markedly improves data reliability.

Built upon the BMPDS corpus, \modelname{} directly integrates our visibility-aware attention mechanism into the diffusion transformer architecture. This mechanism is achieved mainly via an operation inspired by NeRF~\citep{nerf} termed Volumetric Weight Regulation, which explicitly constructs attention masks based on the likelihood of visibility derived from volumetric rendering principles. Unlike prevalent global-level attention, our visibility-aware attention mechanism effectively decouples instance-level spatial embeddings from semantic identity features. This design enables independent and accurate placement of each subject by explicitly modeling occlusion and perspective relationships, and crucially introduces no extra train-time parameters or inference overhead.
Restricting reference features to user-specified regions further enhances identity fidelity and spatial precision, unlocking new applications of controllable image customization (Figure~\ref{fig:fig1_1}).

Our contributions are summarized as follows:
\begin{itemize}
    
    \item  We present BMPDS, the first large-scale automatic data‐synthesis framework for high-fidelity, position-annotated multi‐subject pairs, filling a critical gap in spatial supervision.

    \item We propose \modelname{}, a lightweight yet highly effective layout-aware diffusion framework. It integrates a novel visibility-aware attention mechanism into the diffusion transformer architecture. This design explicitly decouples spatial layout from semantic identity, enabling unprecedentedly accurate multi-subject placement and occlusion-aware rendering without additional training parameters or inference overhead.
    
    
    \item Extensive experimental results demonstrate that our approach delivers state-of-the-art performance on subject-driven customization benchmarks, achieving the highest spatial precision and identity consistency to date.
\end{itemize}

\begin{figure*}[ht]
    \centering
    \includegraphics[width=1\linewidth]{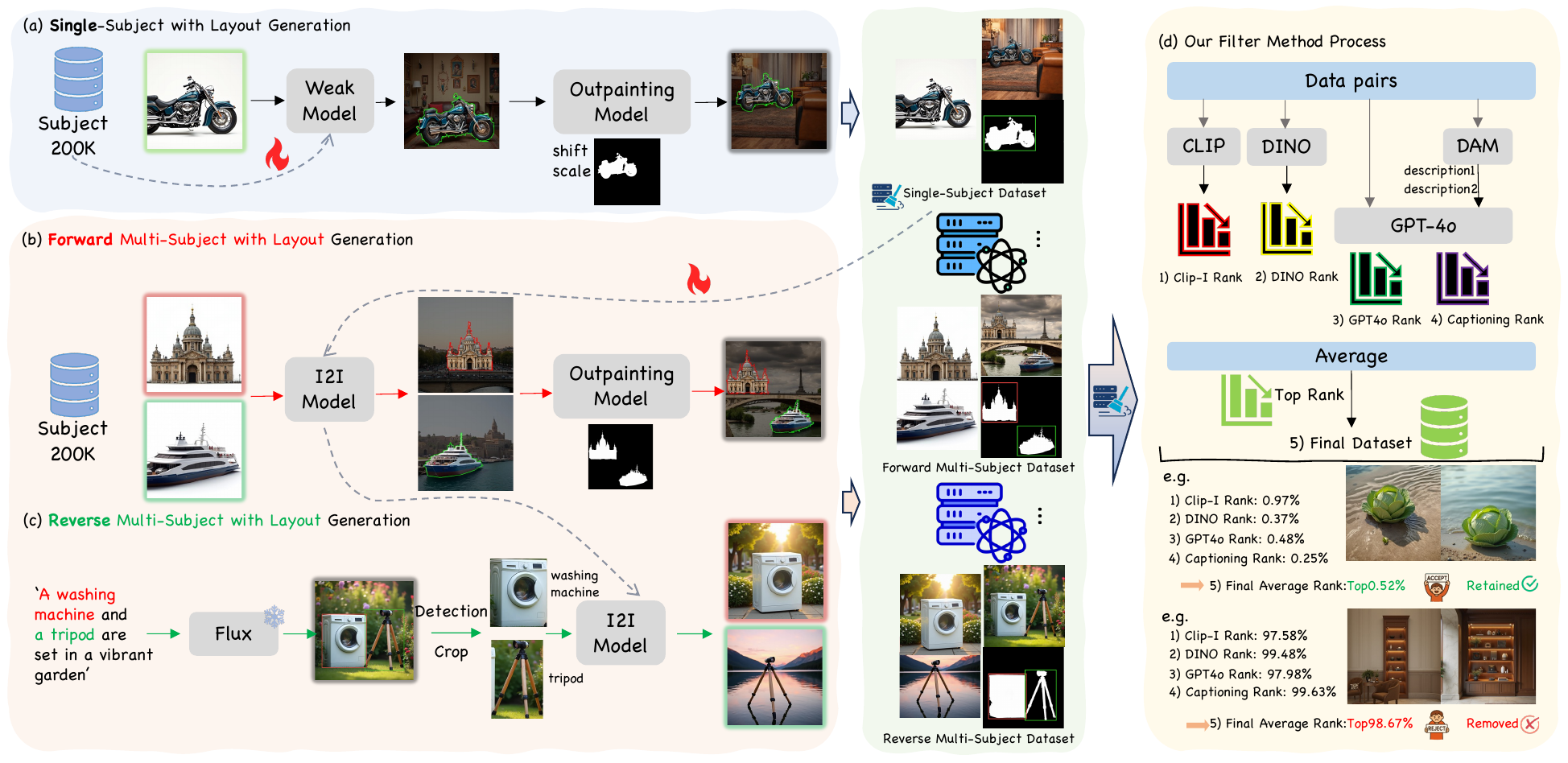}
    \caption{Bidirectional Multi-dimensional Perception Data Synthesis framework. \textbf{(a)} We use Subject200K to train a weak model. \textbf{(b)} Forward generation of multi-subject data pairs.  \textbf{(c)} Reverse generation of multi-subject data pairs. \textbf{(d)} We utilize MLLMs to filter out our data pairs.}
    \label{fig:data_pipeline}
\end{figure*}
\begin{figure*}[ht]
    \centering
    \includegraphics[width=0.9\linewidth]{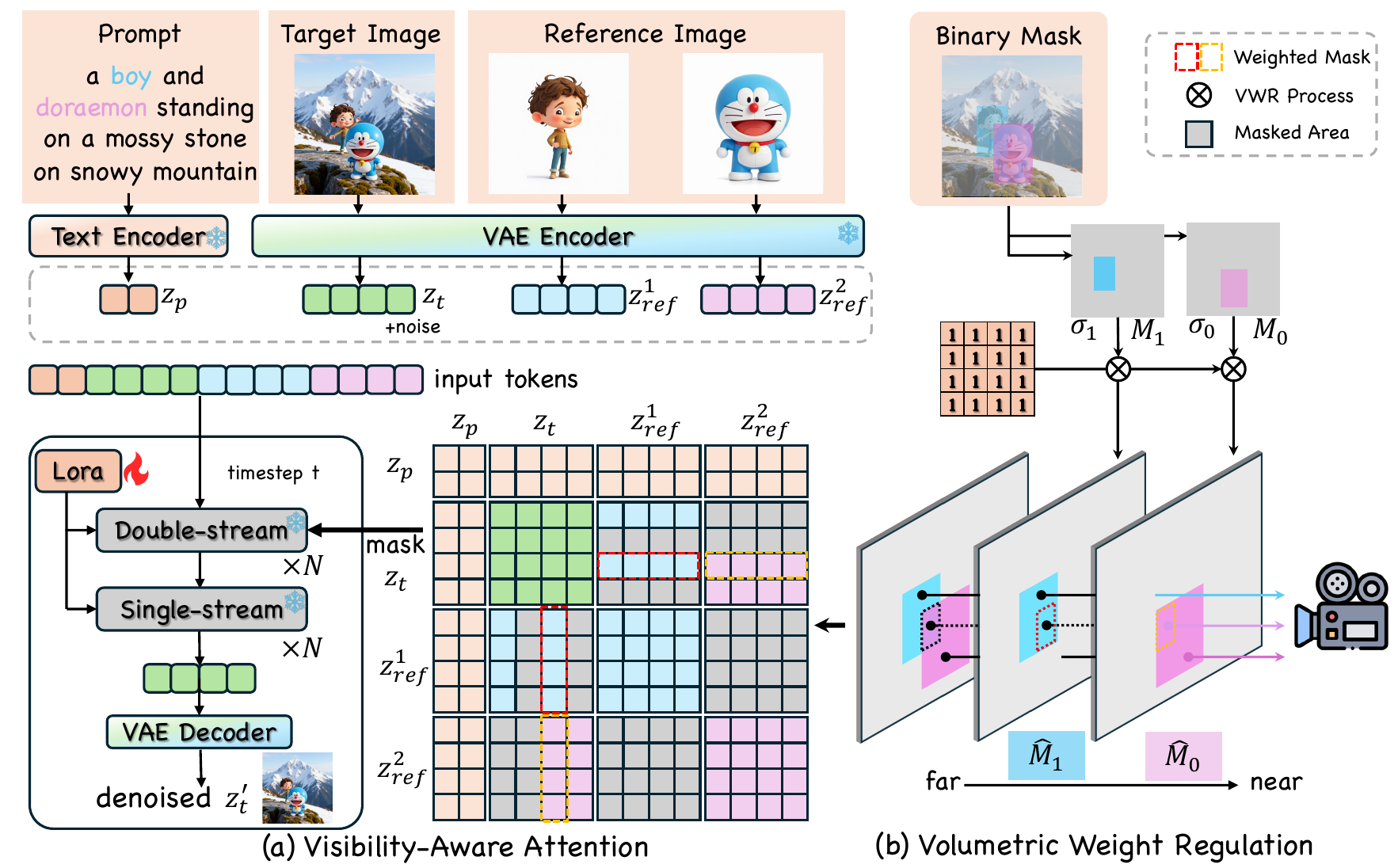}
    \caption{The overall framework of \modelname{}.\textbf{(a)} Reference images and prompts are encoded and concatenated with the latent embeddings $z_{t}$, then the whole token sequence is passed to DiT. Each reference image $z_{ref}^{i}$ is only visible for the specific area of latent noise $z_{t}$ in the attention map. \textbf{(b)} The objects' binary mask and the semantic density $\sigma$ are used by VWR to calculate the weighted mask. The dashed boxes in the ray diagram represent the overlapping regions, where the same colors correspond to the weights in the mask.  }
    \label{fig:pipeline}
\end{figure*}

\section{Related Work}

\subsection{Subject-driven Generation}

In addition to using text prompts for conditional image generation, current diffusion models~\citep{ELITE, li2023blip, huang2024realcustom, ip-adapter, groundingbooth, fastcomposer, MultiDiffusion, feng2025personalize} support reference image input to achieve preservation of subject identity. Dreambooth~\citep{dreambooth} and LoRa~\citep{lora} control the generation of diffusion models through fine-tuning on the specific subject. Recently, Diffisuion-Transformers-based subject-driven models~\citep{uno, omini_control,omini_control2, dreamo, xiao2025omnigen,flux-kontext} further advance subject-driven generation. They introduce reference images as context information through token concatenation to ensure the consistency of objects during generation. 


\subsection{Position Controllable Generation}
Some works~\cite{instancediffusion,regional_flux,wang2024compositional,jimenez2023mixture,MultiDiffusion,shi2025layoutcot,LaRender} have attempted to generate with precise layout control.  Gligen~\citep{gligen} encodes Fourier embedding as grounding tokens to inject position information. MS-diffusion~\cite{ms-diffusion} utilizes a grounding resampler correlating visual information with specific entities and spatial constraints. However, they still suffer from inconsistencies in vision and position, which hinder further application.

\section{Method}

\subsection{\datafiltname ~(BMPDS)}

Position-controllable image-driven customization requires high-fidelity paired data, featuring prominent subjects and high resolution with layout control signals. However, existing open-source datasets such as Subject200K~\citep{omini_control} generate paired data using diptych images, which leads to object inconsistency issues and are limited by low resolution and the lack of positional information. 
We introduce \textbf{BMPDS}, a Bidirectional Multi-dimensional Perception Data Synthesis framework to tackle these limitations. We adopt a hierarchical generation-and-selection strategy to progressively improve data quality, gradually introducing single-image and multi-image pairs with spatial control information during generation. The overall framework is depicted in Figure~\ref{fig:data_pipeline}.



\subsubsection{Customized Data Paired Synthesis}

We divide the automated data generation process into three stages.
\textbf{(1)} Inspired by UNO, we first train a weak model for image customization tasks using Subject200K dataset. The generated results are then segmented and passed through a Flux-Outpainting model with random placement to inject spatial control information. The paired data obtained in this stage is filtered by a filter to ensure high fidelity. These data are then used to train the PositionIC-Single model.
\textbf{(2)} In the second stage, we perform forward generation of multi-image data pairs. We independently process Subject200K samples via PositionIC-Single, then randomly pair and position the output as input to the Flux-Outpainting model, thereby obtaining multi-subject paired data.
\textbf{(3)} To enhance data diversity and improve generalization, we reverse the above process in the third stage. We first use Large Language Models (LLMs) to generate text descriptions containing multiple subjects, then employ the Flux model to create high-resolution images. Objects are detected and cropped from these results, individually processed by the PositionIC-Single, resulting in high-resolution multi-subject paired data.

Through bidirectional data synthesis, we construct \datasetname{}400K dataset. However, we need extra filtering processes to improve data quality as the dataset is still suffering from noise.


\subsubsection{Multi-dimensional Perception Data Filter}\label{filter_descrip}


Previous studies~\citep{cao2024mmfuser,he2025analyzing} have shown that MLLMs have limited capability in recognizing fine-grained details in images. Instead of directly feeding image pairs into the MLLMs for filtering, we establish a reliable system to achieve more accurate and efficient data filtering. 
Specifically, we divide the filtering process into three levels based on granularity. 
\textbf{Firstly}, we segment the subjects from data pairs and utilize CLIP-I~\citep{CLIP} and DINO~\citep{DINO} scores $s_v$ to filter out images with significantly lower consistency.
\textbf{Subsequently}, we pass two subjects' images through MLLMs (e.g., GPT-4o), which directly gives a similarity score $s_{vlm}$ based on shape, color and so on.
\textbf{Lastly}, we employ Describe Anything Model (DAM)~\citep{describe_anything}, a description expert to obtain detailed textual descriptions for each subject. Given these textual description pairs, we instruct GPT-4o as a judge to autonomously select comparative features(e.g., color, shape) and assign multi-dimensional similarity scores $s_{ds}$. 

Then we calculate the ranking separately and average them for each image pair to obtain final ranks:
\begin{align}\label{filter_score}
rank = avg(r(s_v),r(s_{vlm}),r(s_{ds})),
\end{align}

where $r(\cdot)$ denotes rank. With a lower rank indicating a higher subject similarity.

We apply the filter on \datasetname{}400K to rank the pairs and filter out inconsistent pairs. The filtered dataset \datasetname{}98k consists of 44k single-subject pairs and 54k multi-subject pairs. More examples of our \datasetname{}98K are depicted in the appendix.

\subsection{PositionIC }\label{positionic}


\begin{figure*}[ht]
    \centering
    \includegraphics[width=0.9\linewidth]{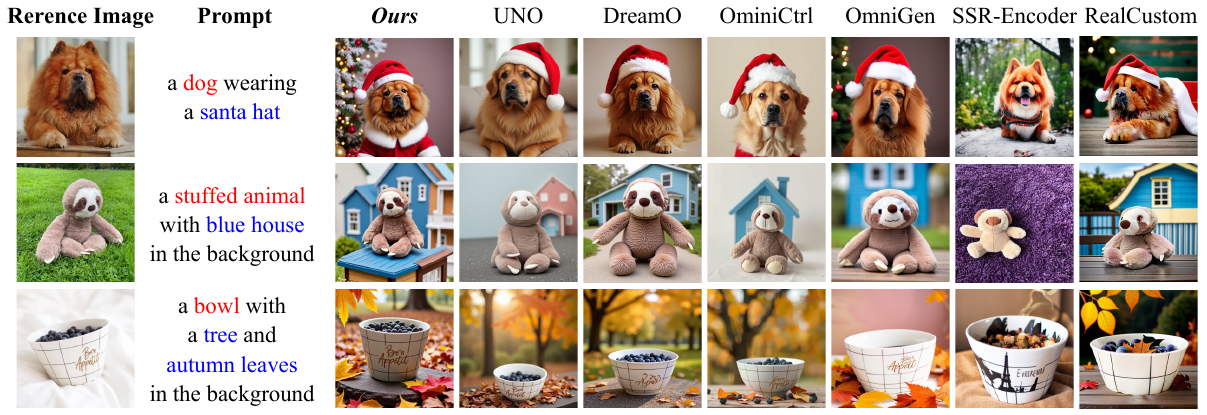}
    \caption{Qualitative comparison of single-subject generation with different methods on DreamBench. }
    \label{fig:single}
\end{figure*}

\begin{figure*}[h]
    \centering
    \includegraphics[width=0.9\linewidth]{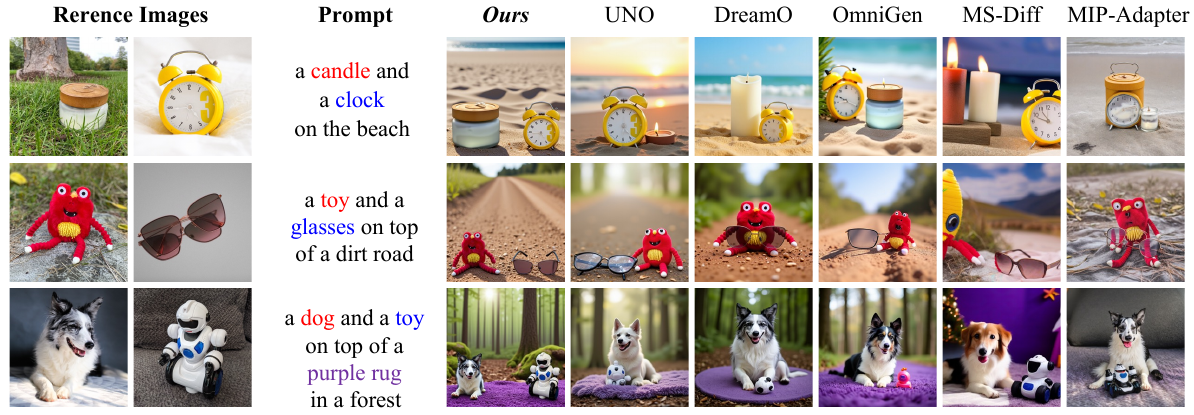}
    \caption{Qualitative comparison of multi-subject generation with different methods on DreamBench. We adopt a fixed bounding box (e.g., bottom left and
bottom right) for generation.}
    \label{fig:multi}
\end{figure*}


When extending text-to-image models to image-to-image tasks, the fidelity of the generated object largely depends on the magnitude of the attention between the corresponding regions in the target image and the reference image. We refer to this as attention accumulation.
As shown in Figure~\ref{fig:pipeline}, the attention map can be divided into four areas: text-text self attention, image-image self attention, text-image and image-text cross attention. If extending to single or multi-subject custom generation via concatenating method like~\citep{uno,omini_control, flux-kontext,eligen}, the attention map will expand to nearly three times its original size, making it harder for the model to focus on the corresponding regions.

To achieve effective attention accumulation, 
we propose \textbf{\nerfmodelname{}}, which utilizes volume rendering(VR) to explicitly define the area that can be focused on for each reference image.
Previous works~\citep{eligen, regional_flux,zhang2025creatidesign,LaRender} explore the effort of restricting the attention horizon between special words and noise, we extend it to subject-driven generation task to unlock the positional control ability of diffusion transformers (DiT).
The overall mechanism is shown in Figure~\ref{fig:pipeline}.

\subsubsection{Volume Rendering(VR)}
The weighted mask utilizes VR~\citep{volume_render} to handle the occlusion relationships between individual objects. Similar to NeRF~\citep{nerf}, the expected color $C(r)$  observed by the camera along the distance from the far bound $t_f$ to the near bound $t_n$ is:
\begin{align}\label{eq:nerf}
& C(r)=\int_{t_n}^{t_f}T(t)\sigma(r(t))c(r(t),d)dt,\notag \\
& T(t)=exp(-\int_{t_n}^{t}\sigma(r(s))ds)
\end{align}
where the camera ray $r(t)=o+td$, $\sigma(x)$ is volume density. We utilize the quadrature rule to obtain a discrete estimate of $C(\mathbf{r})$:
\begin{align}\label{eq:discrete_nerf}
& \hat{C}(r)=\sum_{i=1}^{N}T_i(1-exp(-\sigma_i\delta_i))c_i, \notag \\ 
& T_i=exp(-\sum_{j=1}^{i-1}\sigma_i\delta_i)
\end{align}
where $N$ denotes the spatial resolution for the distance range from near to far bounds. $\delta_i$ is the distance between adjacent samples.


\subsubsection{Volumetric Weight Regulation}
Similar to the 3D rendering process, controlling the generation position of individual objects on a 2D canvas can be viewed as a composite image captured by an orthographic virtual camera. 
To precisely distinguish the occlusion relationships between objects, we leverage VR to modulate a physically plausible semantic mask.
Given the foreground mask $M_i$ and the sample distance $\delta$, the volumetric weight $\hat{M_n}$ can be formulate as:
\begin{align}\label{eq:our_nerf}
& \hat{M}_{n}=exp(-\sum_{j=1}^{n-1}\sigma_i\delta)(1-exp(-\sigma_n\delta))M_n\hat{M}_{n-1} 
\end{align}
unlike in ~\cref{eq:discrete_nerf}, $\hat{M_n}$ is defined as the volumetric weight for each distance level, which is utilized to represent the degree of attention the image pays to each region of the reference image during attention calculation. Specifically, we utilize a semantic density $\sigma_i$ instead of the volume density to represent the semantic interaction between objects. Setting of $\sigma_i$ can be found in appendix.  

\subsubsection{\nerfmodelname{}}

As shown in Figure~\ref{fig:pipeline}, we first encode the reference images $I_{r}^{i}$ to the latent space via VAE $\epsilon(\cdot)$. Encoded results $z_{ref}$ are then concatenated with noise latents $z_{t}$ and text embedding $z_{p}$. This process can be formulated as:
\begin{align}\label{eql.z}
& z_{ref} = [\epsilon(I_{r}^{1}),\epsilon(I_{r}^{2}),...,\epsilon(I_{r}^{N})], \notag\\
& z = Concatenate(z_{p},z_{t}, z_{ref}),
\end{align}


where $N$ is the number of reference images and $z$ denotes the input tokens to the DiT model.


Since the global text prompt contains information about the entire image, generating a reference object in a specific region requires focused and singular attention. They only need to pay attention to the corresponding region while ignoring other irrelevant tokens. 
Therefore, we use a VWR-based attention mask, which we call visibility-aware attention(VAA), to shield the regions that each token should not directly focus on. 
As shown in Figure~\ref{fig:pipeline}, each reference image $z_{ref}$ has a restricted attention horizon, which blocks the attention between itself and other reference images. Moreover, only the area within the bounding box in the noise is visible for the specific reference image. The mechanism of VAA mask $M$ can be formulated as:
\begin{align}\label{attention_horizon_mask}
& M(z_{ref}^i,z_{ref}^j)=0, i\neq j, \notag\\
& M(z_{ref}^i,z_t^n)=\hat{M}_{i}, \notag\\
& M(other) = 1,
\end{align}
where $\hat{M}_{i}$ is volumetric weight of $i^{th}$ reference subject.
Thus, the computation of attention is derived as:
\begin{equation}\label{eql_attention}
Attention = Softmax(\frac{QK^{T}}{\sqrt{d}}+\log{M})\cdot V
\end{equation}







\subsection{\benchname{}}

Most existing customized image evaluation (e.g., DreamBench) lacks explicit spatial position annotations. Thus, there is no universal data benchmark for evaluating subject-driven methods with position control. To address this gap, we propose \benchname{}, a benchmark to evaluate subject consistency and position accuracy simultaneously. 

We manually select 252 single-subject samples and 296 multi-subject samples in the benchmark, where the object bounding boxes conform to standard proportions and include challenging positional relationships. 



\begin{figure*}
    \centering
    \includegraphics[width=0.9\linewidth]{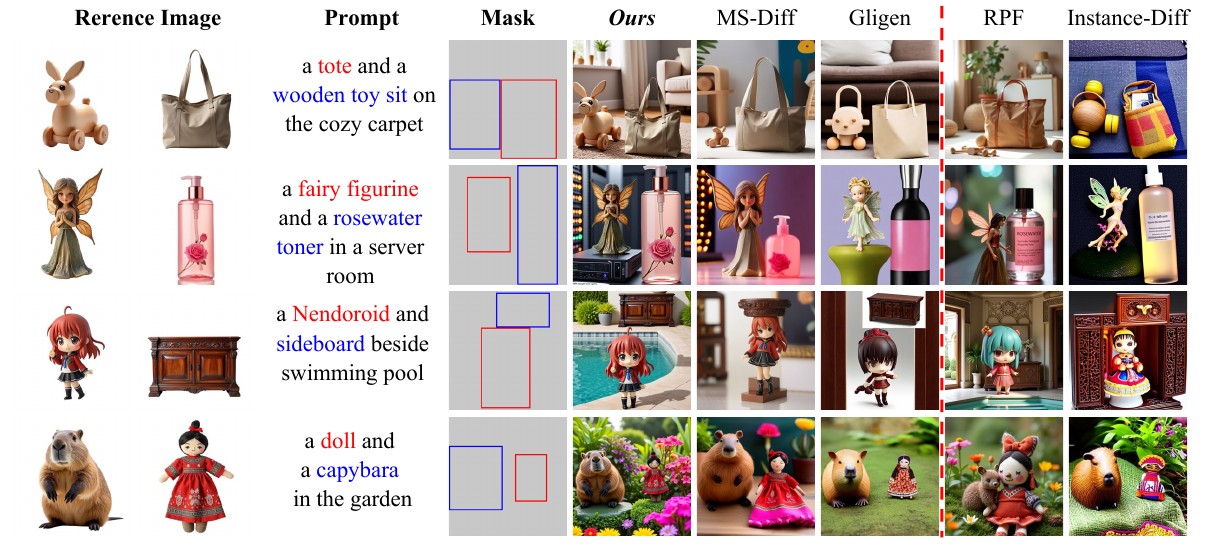}
    \caption{Qualitative comparison of position control generation with different methods on \benchname{}. MS-Diff and RPF denote MS-Diffusion and Regional Prompting Flux respectively. MS-Diff and Gligen are existing position-controllable customization methods; RPF and Instance-Diff are position-only controllable methods.}
    \label{fig:layout}
\end{figure*}

\section{Experiments}
\subsection{Implementations}
\subsubsection{Training Detail}

Following UNO, we first initialize the model using FLUX.1 dev and apply UnoPE to extend the position embedding of the reference images to the non-overlapping area diagonally. We train a LoRA at the rank of 512 on 8 NVIDIA A100 GPUs and set the total batch size of 128. The learning rate is set to ${10}^{-5}$ with cosine warm up. In the first stage, we train the single-subject model on 44k single-subject pairs for 10k steps. We then continue our training on 54k multi-subject pairs for 8k steps, which extends the multi-subject generation capability to the model obtained in the first stage.



\begin{table}[ht]
  \centering
  \begin{tabular}{lccc}
    \toprule
     Method &  CLIP-I$\uparrow$ & CLIP-T$\uparrow$ & DINO$\uparrow$ \\
    
    \midrule
    Dreambooth 
    & 0.776 & 0.215 & 0.679\\
    BLIP-Diffusion
    & 0.787 & 0.234 & 0.742 \\
    ELITE
    & 0.803 & 0.235 & 0.723  \\
    SSR-Encoder
    & 0.797 & 0.206 & 0.725  \\
    RealCustom
    & 0.783 & 0.242 & 0.765 \\
    MS-Diffusion
    & 0.808 & 0.242 & 0.791  \\
    OmniGen
    & 0.791 & 0.267 & 0.751 \\
    DreamO 
    & 0.835 & 0.258 & 0.802 \\
    OminiControl 
    & 0.805 & \underline{0.268} & 0.735 \\
    UNO 
    & \underline{0.840} & 0.253 & \underline{0.814} \\
    \midrule
    \modelname$({Ours})$ & \textbf{0.846} & \textbf{0.269} & \textbf{0.823} \\
    \bottomrule
  \end{tabular}
  \caption{Evaluation on DreamBench for single-subject driven generation. The \textbf{bold} value is the highest and the \underline{underlined} value is the second.}
  \label{metric}
\end{table}

\begin{table}[h]
  \centering
  \begin{tabular}{lccc}
    \toprule
     Method &  CLIP-I$\uparrow$ & CLIP-T$\uparrow$ & DINO$\uparrow$ \\
    
    \midrule
    BLIP-Diffusion 
    & 0.703 & 0.212 & 0.541 \\
    MIP-Adapter 
    & 0.752 & 0.254 & 0.657  \\
    MS-Diffusion 
    & 0.772 & 0.261 & 0.683  \\
    DreamO 
    & 0.779 & 0.273 & 0.698 \\
    UNO 
    & \underline{0.781} & \underline{0.279} & \underline{0.707} \\
    OmniGen 
    & 0.749 & \textbf{0.291} & 0.668 \\
    \midrule
    \modelname$({Ours})$ & \textbf{0.819} & \underline{0.279} & \textbf{0.771} \\
    \bottomrule
  \end{tabular}
  \caption{Evaluation on DreamBench for multi-subject driven generation.}
  \label{metric2}
\end{table}

\subsubsection{Evaluation Metrics}

For subject-generation tasks, we evaluate subject similarity using CLIP-I, DINO-I on Dreambench~\citep{dreambooth}. For text fidelity, we calculate CLIP-T scores, which measure cosine similarity between the text embedding and image embedding from CLIP. 

For the position-guided task, we evaluate different methods on our proposed \benchname{}.
We use Vision-R1~\citep{visionr1} to determine the bounding box of the subject and calculate mIoU and AP scores with the label.

\begin{table*}[h]
  \centering
  \begin{tabular}{lcc|cc}
    \toprule
    \multirow{2}{*}{Method} &  \multicolumn{2}{c}{Single-Subject} & \multicolumn{2}{c}{Multi-Subject} \\ 
     \cmidrule{2-5}
      &  IoU$\uparrow$ & $AP\uparrow$ / $AP_{50}\uparrow$ / $AP_{70}\uparrow$ & mIoU$\uparrow$ & $AP\uparrow$ / $AP_{50}\uparrow$ / $AP_{70}\uparrow$\\
    
    \midrule
    RPF~\citep{regional_flux} 
    & 0.341 & 0.015 / 0.063 / 0.007 & 0.369 & 0.070 / 0.002 / 0.011  \\
    MS-Diffusion~\citep{ms-diffusion}
    & 0.501 & 0.097 / 0.329 / 0.075 & 0.421 & 0.028 / 0.146 / 0.005  \\
    Instance-Diffusion~\citep{instancediffusion}
    & 0.789 & 0.593 / 0.683 / 0.632 & 0.799 & 0.497 / 0.699 / 0.546 \\
    Gligen~\citep{gligen}
    & \underline{0.808} & \textbf{0.632} / \underline{0.865} / \textbf{0.811} & \underline{0.825} & \underline{0.628} / \underline{0.858} / \underline{0.811} \\
    \midrule
    \modelname$({Ours})$ & \textbf{0.828} & \underline{0.628} / \textbf{0.904} / \underline{0.761} & \textbf{0.860} & \textbf{0.701} / \textbf{0.939} / \textbf{0.853} \\
    \bottomrule
  \end{tabular}
  \caption{Quantitative results of controllable spacial generation on \benchname{}. }
  \label{layout_metric}
\end{table*}

\subsection{Qualitative Result}
We visualize the comparison results with the current state-of-the-art methods in Figure~\ref{fig:single}. Overall, our method surpasses all current methods in terms of visual effects. It can be seen that our method can still effectively follow the prompt while maintaining subject consistency, demonstrating higher text fidelity. Other methods either fail to follow complex instructions or cannot maintain consistency. (e.g., UNO and DreamO cannot reproduce the dog face, while SSR-Encoder fails to add the Santa hat.)
The results of the third row also reveal that \modelname{} has a great capability on patterns and text.
Furthermore, \modelname{} consistently produces images with a higher degree of naturalness and visual plausibility. More examples can be found in Appendix.

Figure~\ref{fig:multi} shows the comparison of multi-subject generation. To control variables, the images generated by our method adopt a fixed bounding box (e.g., bottom left and bottom right). In a more difficult multi-subject scenario, \modelname{} can still maintain high subject similarity and follow the given text prompt, whereas results from most other methods fail to preserve consistency for each subject and even ignore certain subjects.

We further evaluate the positional control capability of existing methods on \benchname{} via randomly generating the bounding boxes. As shown in Figure~\ref{fig:layout}, \modelname{} accurately generates subjects that fully occupy the bounding box without damaging their features. More importantly, our flux-based approach significantly outperforms others in terms of image aesthetic quality and the logical coherence of multi-subject compositions.


\begin{figure}[ht]
    \centering
\includegraphics[width=1\linewidth]{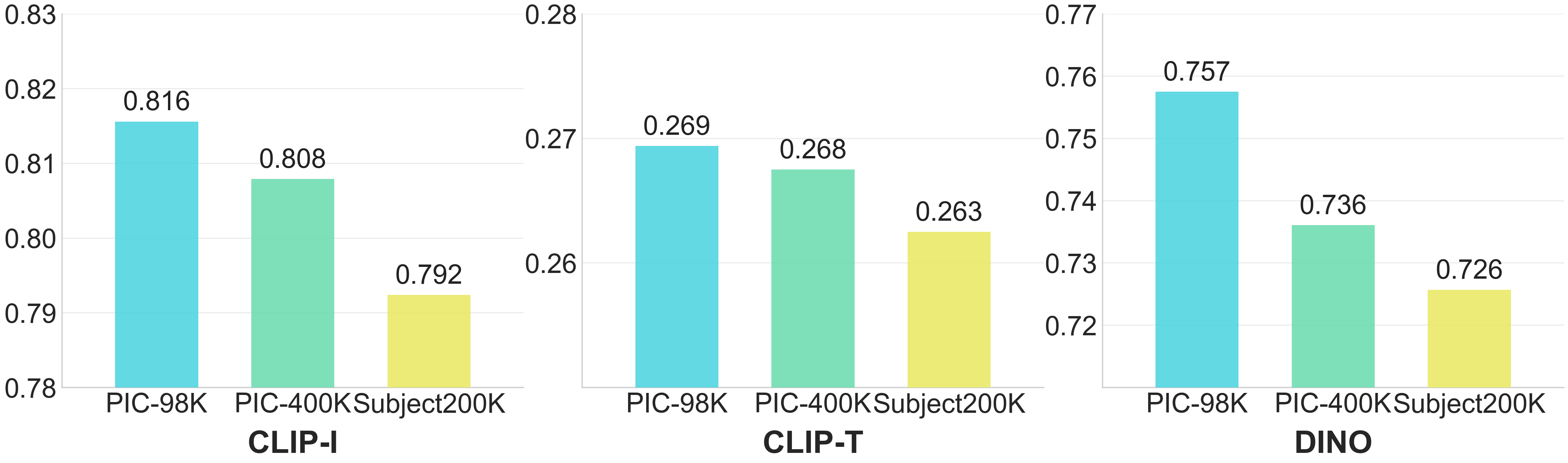}
    \caption{Ablation study of data filter. We train our model on \datasetname{98K}, \datasetname{400K} and Subject200K respectively.}
    \label{fig:data_compare}
    \vspace{-2mm}
\end{figure}

\subsection{Quantitative Evaluations}
\subsubsection{Subject-driven Analyses}
Specifically, we discover that different object sizes can lead to inaccurate scores, which will be detailed in Appendix. 

We compare our proposed \modelname{} with several leading methods on DreamBench for both single-subject and multi-subject. As presented in Table~\ref{metric} and Table~\ref{metric2}, \modelname{} achieves the highest scores on both CLIP-I and DINO of 0.846 and 0.823 on single-subject, 0.819 and 0.771 on multi-subject, respectively. \modelname{} also has competitive CLIP-T compared to existing methods. The evaluation results indicate that \modelname{} has remarkable performance on subject consistency and text fidelity.

\begin{figure}
    \centering
    \includegraphics[width=1\linewidth]{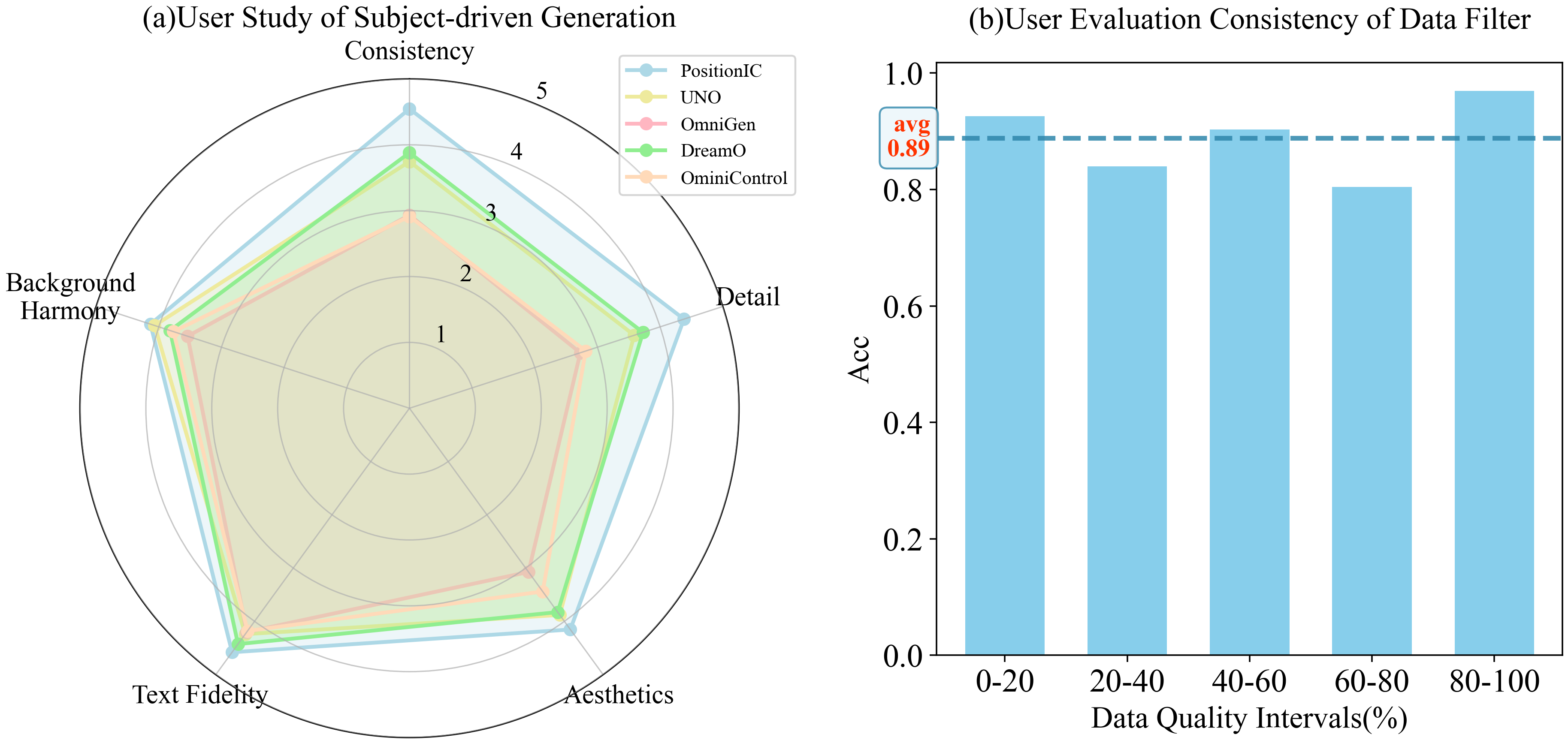}
    \caption{User study for subject-driven generation and data filter. \textbf{(a)} User evaluation on Dreambench. \textbf{(b)} Filtering consistency of our data filter compared with human in different data quality intervals (from good to bad).}
    \label{fig:user_study}
    \vspace{-5mm}
\end{figure}

\subsubsection{Controllable Spacial Generation}


Table~\ref{layout_metric} presents the results of spatial control evaluation. \modelname{} achieves superior performance in both single-subject and multi-subject position control. For single-subject position control, \modelname{} has the highest IoU of 0.828 across all methods and has competitive $AP$ scores compared with Gligen. For multi-subject evaluation, \modelname{} achieves the highest scores in both AP and IoU scores, demonstrating a significant advantage over existing methods.

\subsubsection{User Study}
We invite evaluators for an extensive user study. For subject-driven generation, we randomly selected 500 images from the results on DreamBench for manual evaluation. Six users scored the results from five dimensions, each ranging from 0 to 5, and the average score was taken. The result presented in Figure~\ref{fig:user_study} (a) shows that our method reaches the highest capability to preserve image features and details while maintaining competitive text adherence.

To evaluate the consistency between human annotators and the data filter in BMPDS, we use percentage agreement metric (denoted as Acc.), comparing the filter's output against human-generated annotations. As shown in Figure~\ref{fig:user_study} (b), our filter has an average consistency of 0.89 with human annotators.


\subsection{Ablation Study}
In this section we show the ablation study of our key modules, including data quality and Volumetric Weight Regulation(VWR). More ablation results are presented in the appendix.

\subsubsection{Impact of data filter.} 
Results in Figure~\ref{fig:data_compare} illustrate the advanced fidelity of BMPDS. We directly trained with \datasetname{}98K, \datasetname{}400K and the Subject200K dataset without injecting position control information. \datasetname{}400K achieves significantly higher scores than Subject200K, and the filtered data \datasetname{}98K achieves the highest scores overall, which demonstrates the remarkable efficiency of our data synthesis and filtering pipeline. 


 \begin{figure}[h]
     \centering
     \includegraphics[width=1\linewidth]{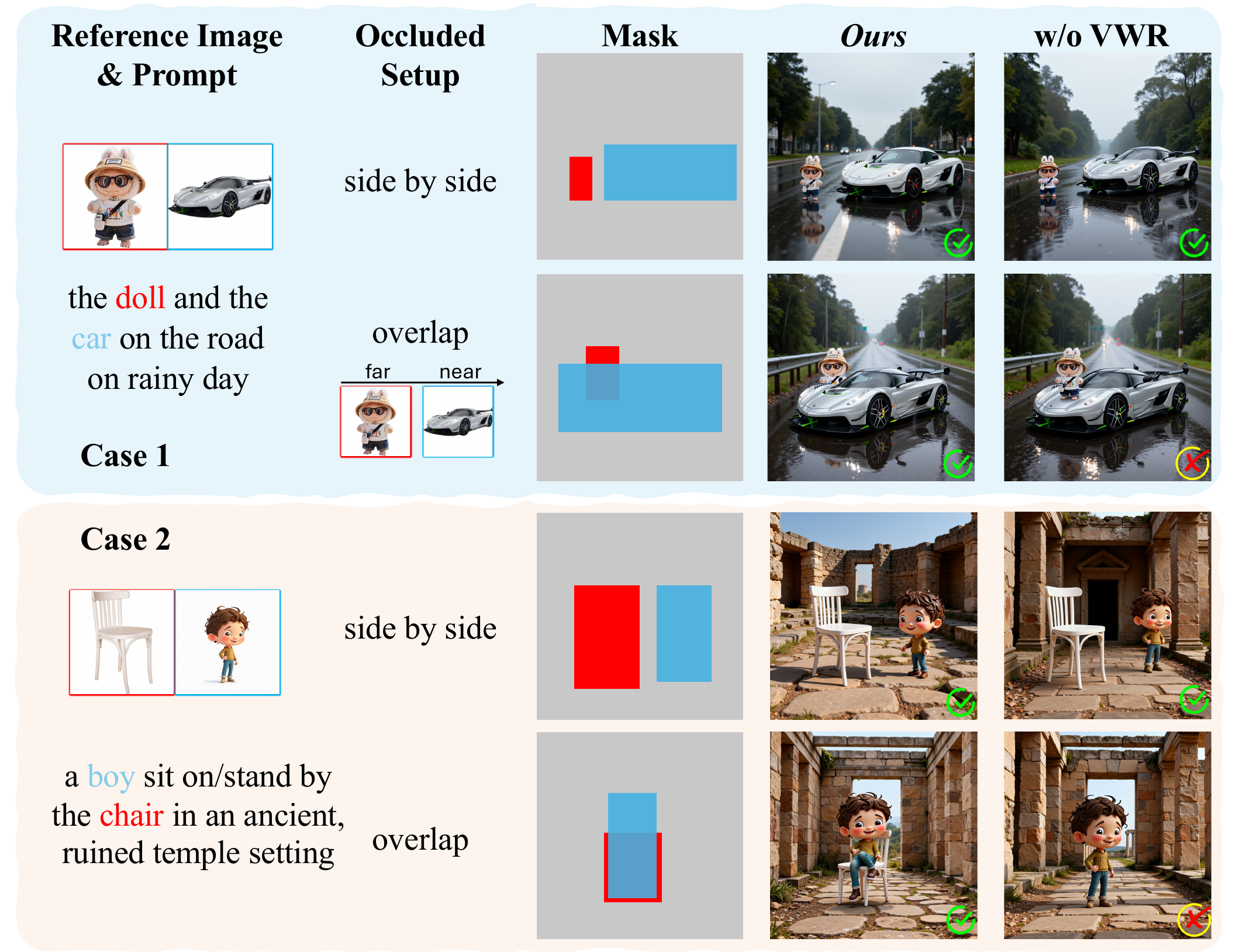}
     \caption{Impact of VWR. In the presence of overlaps, VWR is capable of correctly handling the spatial relationships between objects. We set the attention value in overlap region as 1 to disable VWR.}
     \label{fig:ablation_nerf}
 \end{figure}


\subsubsection{Abation results of Volumetric Weight Regulation(VWR)}
As shown in~\cref{fig:ablation_nerf}, in Case 1 overlap setting, the doll is instructed to be placed behind the sports car. However, the model fails to correctly understand their spatial relationship without VWR.
In Case 2, VWR is able to prevent the generation of missing objects caused by concept confusion, even when multiple objects overlap.

\section{Conclusion}
In this work, we present \modelname{}, an innovative framework capable of customizing multiple subjects with precise position control. \modelname{} decouples layout signal from subject feature without introducing additional parameters and training cost. Additionally, we carefully design an automatic data curation framework to obtain high-fidelity paired data. We adopt bidirectional generation and present a multidimensional perception filter to improve object consistency in acquired data. Extensive experiments demonstrate that \modelname{} performs high-quality generation in both single-subject and multi-subject consistency, as well as in controllable subject positioning. We hope our work can advance the development of controllable image customization.

{
    \small
    \bibliographystyle{ieeenat_fullname}
    \bibliography{main}
}

\clearpage
\setcounter{page}{1}
\maketitlesupplementary


\begin{figure}[t]
    \centering
    \includegraphics[width=1\linewidth]{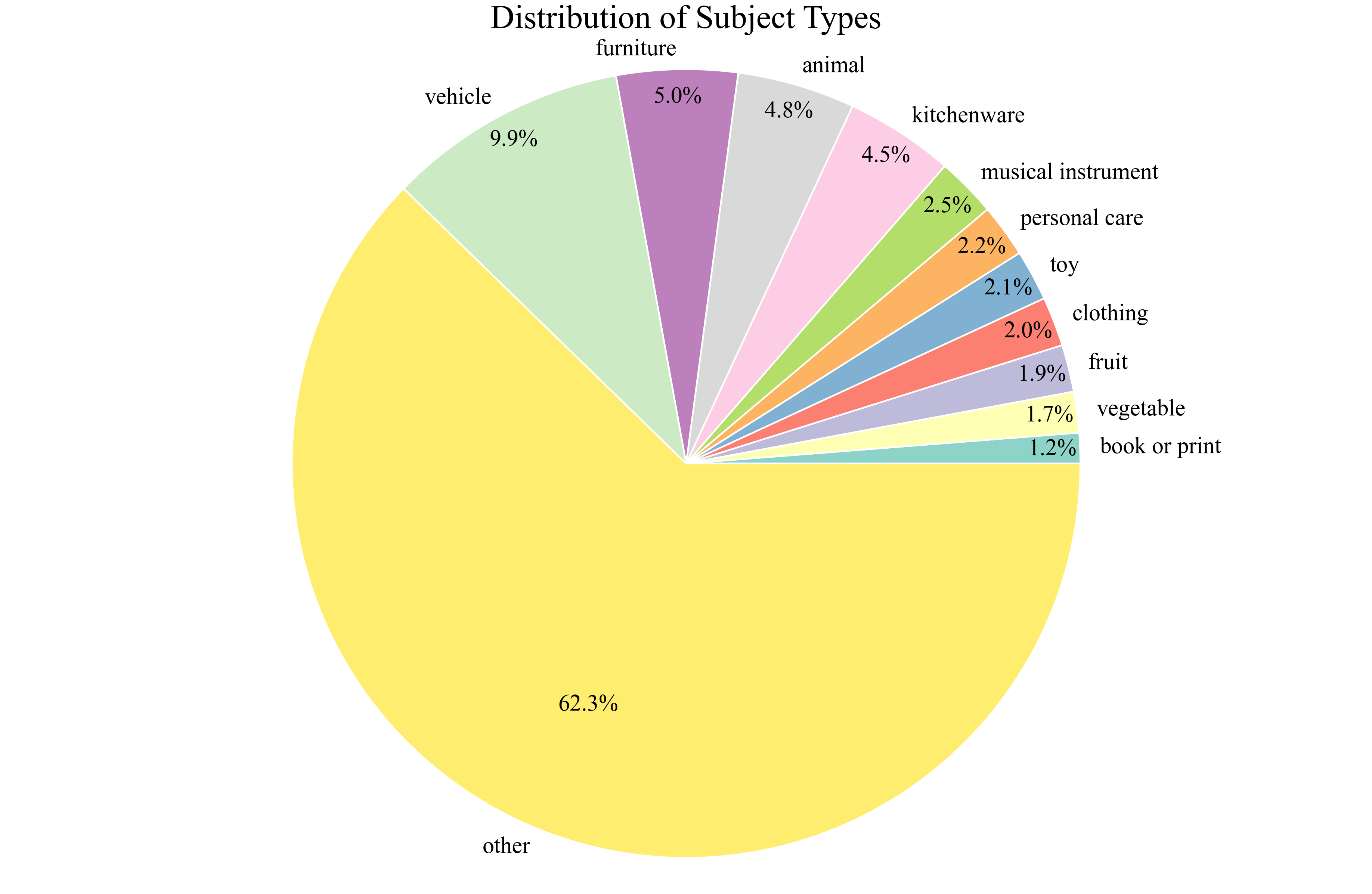}
    \caption{Category distribution of PIC-98K.}
    \label{fig:subject_distribution}
\end{figure}

In Section~\ref{sec:bmpds}, we give a detailed description of our data synthesis and filtering pipeline: Bidirectional Multi-dimensional Perception Data Synthesis (BMPDS). We further present a selection of example images. Section~\ref{sec:positionic_bench} introduces the PositionIC-Bench we use for position evaluation.

Section ~\ref{sec:result},\ref{sec:metric} and \ref{sec:ablation} provide a detailed discussion of specific experimental setups and additional ablation studies. Furthermore, more example demonstrations can be found in Figure~\ref{fig:result}.

\section{Bidirectional Multi-dimensional Perception Data Synthesis}\label{sec:bmpds}
\subsection{Detailed Instruction of GPT-4o}
As shown in Figure~\ref{fig:gpt_instruction}, the message given to GPT-4o consists of \textbf{Instruction}, \textbf{Evaluation Metric} and \textbf{Response}. In the part of \textbf{Instruction}, we have defined the input format and evaluation metric dimensions, and required GPT-4o to select no fewer than three features for scoring based on the content of textual description. In the next part of \textbf{Evaluation Metric}, we detail the metric standard and provide GPT-4o examples to evaluate. There are 6 levels, ranging from 0 to 5, representing the similarity between two descriptions regarding the same feature. After that, we prompt GPT-4o to return a dictionary in JSON format containing the subject types and the scores for each feature. If there is no similarity between the two descriptions, the subject is set to "none", indicating that the final score is 0.

Due to the substantial differences in textual descriptions between subjects, it is not feasible to predetermine the feature categories for evaluation. Therefore, we allow the LLMs to select at least three features and assign individual scores to each. The final score is calculated as the average of all feature scores. For descriptions with significant discrepancies, the LLMs are permitted to assign a score of zero to the samples.

Figure~\ref{fig:samples_gpt4o} demonstrates the samples of Multi-dimensional Perception Data Filter. We have highlighted the correlated features in the description. In the first sample, the teddy bear share the same physical characteristics except for their posture, hence earning the highest appearance score and slightly lower posture score. In the second sample, the deer is missing antlers, which resulted in the lowest score on the "antler" feature.

\subsection{Details of PIC-98K Dataset} \label{sec:app_dataset}
We propose PIC-400K utilizing our Bidirectional Multi-dimensional Perception Data Synthesis, an automatic and effective high-consistency data synthesis pipeline. Samples of the filtered data PIC-98K are shown in the Figure~\ref{fig:samples_PIC98k}. BMPDS can synthesize high-fidelity multi-subject images while maintaining high resolution. Against previous works, the position of subjects is controllable and it is randomly placed to train the position control capability of PositionIC.

There are over 9000 subject descriptions in PIC-98K, including multiple categories such as fruits, animals, and transportation vehicles, which basically cover common objects. The distribution of subjects is shown in Figure~\ref{fig:subject_distribution}, vehicles, furniture, animal, and kitchenware constitute a significant proportion, with most difficult-to-classify subjects categorized as "other".

\begin{figure}[t]
    \centering
    \includegraphics[width=1\linewidth]{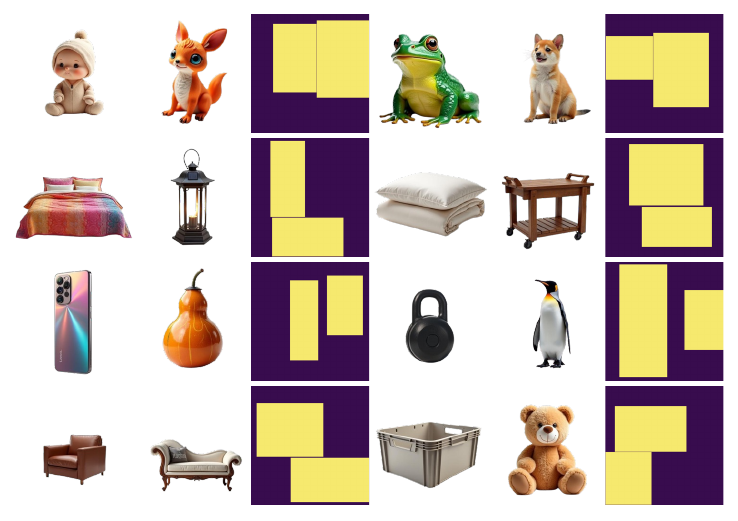}
    \caption{Showcases of PositionIC-Bench.}
    \label{fig:layout_bench}
\end{figure}

\begin{figure*}[ht]
    \centering
    \includegraphics[width=0.9\linewidth]{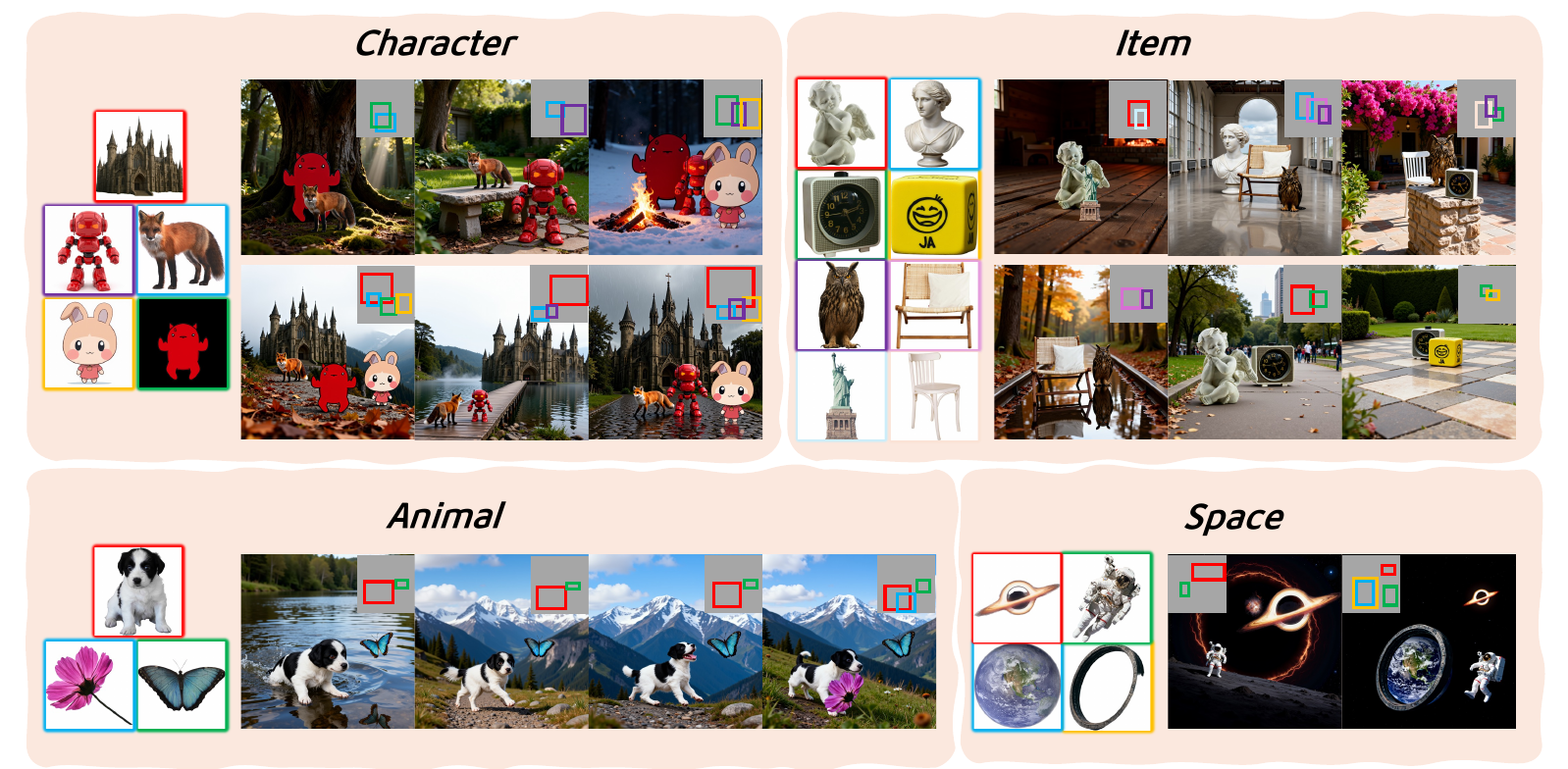}
    \caption{ More results about \modelname{}.}
    \label{fig:result}
\end{figure*}

\section{PositionIC-Bench}\label{sec:positionic_bench}
We manually select 252 single-subject samples and 296 multi-subject samples in the benchmark, where the object bounding boxes conform to standard proportions and include challenging positional relationships. We show some samples of PositionIC-Bench in Figure~\ref{fig:layout_bench}. Our bench includes various subjects such as furniture, animals, plants, and portraits. Not limited to conventional object placement, PositionIC-Bench's bounding boxes have more complex spatial relationships where objects are placed on different planes. At the same time, the bounding boxes of smaller objects is appropriately enlarged to obtain more accurate evaluation scores.

\begin{figure}[t]
    \centering
    \includegraphics[width=1\linewidth]{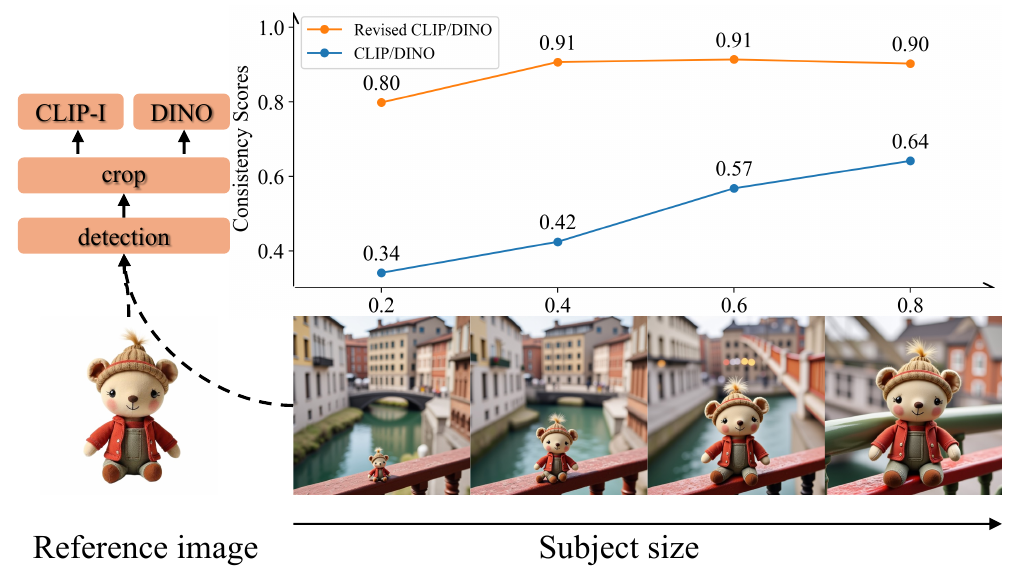}
    \caption{ Inaccuracy of directly using CLIP-I and DINO. The revised score is less affected by the size of the subject which can reflect the subject consistency more authentically.}
    \label{fig:size_compare}
\end{figure}

\section{Training Details and Result}\label{sec:result}
\subsection{Setting of Training}\label{sec:train_set}
The value of the semantic density $\sigma_i$ can be manually set or determined by~\cref{eq:sigma}. In ~\cref{eq:sigma}, we set the semantic density to a geometric progression with a common ratio of $1 + 10\lambda$, which represents that $\sigma_i$ increases incrementally from far to near to reflect the difference between foreground and background objects.

\begin{align}\label{eq:sigma}
\sigma_i = \frac{10\lambda*(1+10\lambda)^i}{(1+10\lambda)^n-1},
\end{align}
where $n$ is the number of reference images and $i=\{0,1,2,...,n-1\}$ represents the object index ordered from far to near. $\lambda$ is a hyperparameter used to control the variation in semantic density. A smaller value of $\lambda$ results in less difference in semantic density between objects. During training, we set it to 5. 

\subsection{More Result}
As shown in Figure~\ref{fig:result}, we downloaded several foreground images from the internet as additional visual demonstrations, covering four categories: Characters, Items, Animals, and Space. In the Characters and Items sections, \modelname{} exhibits high fidelity and is capable of generating images with coherent spatial relationships. In the Animals section, we demonstrate the plausibility of object interactions under different prompts. In the Space section, \modelname{} is able to achieve good results even for difficult examples (e.g., generating a planetary ring surrounding the Earth).

\section{Revised Evaluation Metric}\label{sec:metric}
In this section, we elaborate on the calculation methods for our evaluation metrics. As shown in Figure~\ref{fig:size_compare}, directly using the entire image to compute the CLIP or DINO metrics is unreasonable, as both CLIP and DINO compute the similarity of global image features. To avoid the sensitivity, we crop the subject from original image as source images for evaluation.

\begin{figure}[t]
 \centering
 \includegraphics[width=1\linewidth]{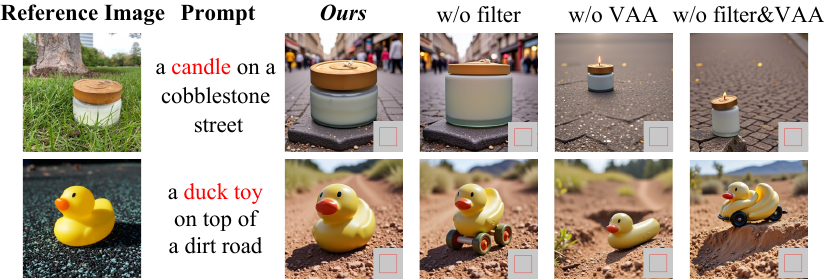}
 \caption{Qualitative results of ablation study. \nerfmodelname{}(VAA) and our filter pipeline are capable of effective position control and consistent customization.}
 \label{fig:ablation_compare}
\end{figure}

\section{More Ablation Study}\label{sec:ablation}
In this section, we conduct more detailed ablation studies, including ablations on the data filter and \nerfmodelname{}(VAA).

\subsection{Impact of \nerfmodelname{}}
As shown in Figure~\ref{ablation_mask}, CLIP-I and DINO scores significantly drop when training without VAA. We infer that incorporating a attention mask allows the model to focus more effectively on features in a smaller region rather than globally, which accelerates the convergence and improves the consistency.

\subsection{Qualitative Results }
As shown in Figure~\ref{fig:ablation_compare}, we visualize the effect of VAA and data filtering. It can be seen that without VAA, the model loses the ability to control position. When using unfiltered data, although the data volume is larger, the generated quality is poorer.

\begin{figure*}[ht]
    \centering
    \includegraphics[width=0.9\linewidth]{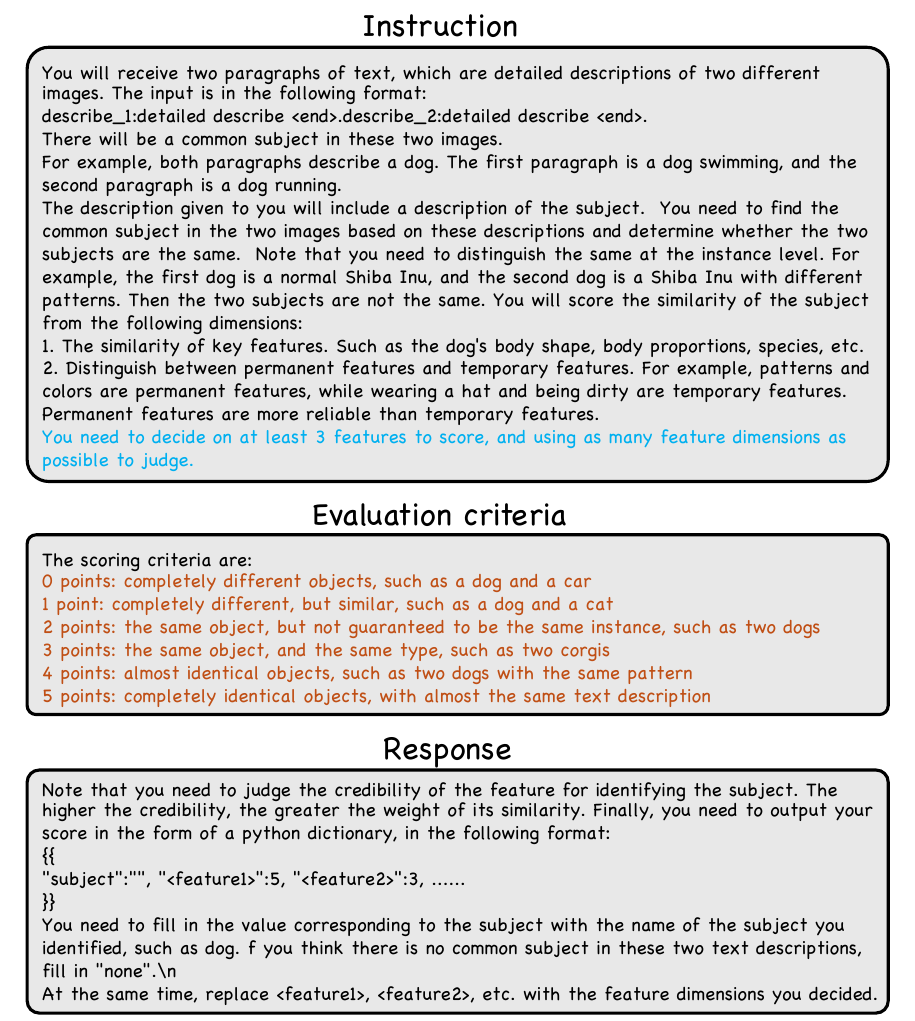}
    \caption{Prompt template of MLLMs in Multi-dimensional Perception Data Filter.}
    \label{fig:gpt_instruction}
\end{figure*}

\begin{table}[ht]
  \centering
  \begin{tabular}{lccc}
    \toprule
     Method &  CLIP-I$\uparrow$ & CLIP-T$\uparrow$ & DINO$\uparrow$ \\
    
    \midrule
    w/o VAA 
    & 0.784 & 0.269 & 0.686  \\
    w/ VAA 
    & 0.846 & 0.269 & 0.823  \\
    \bottomrule
  \end{tabular}
  \caption{Ablation study of VAA. Our model performs better subject fidelity on DreamBench after restricting the attention area of reference images.}
  \label{ablation_mask}
\end{table}


\begin{figure*}[t]
    \centering
    \includegraphics[width=1\linewidth]{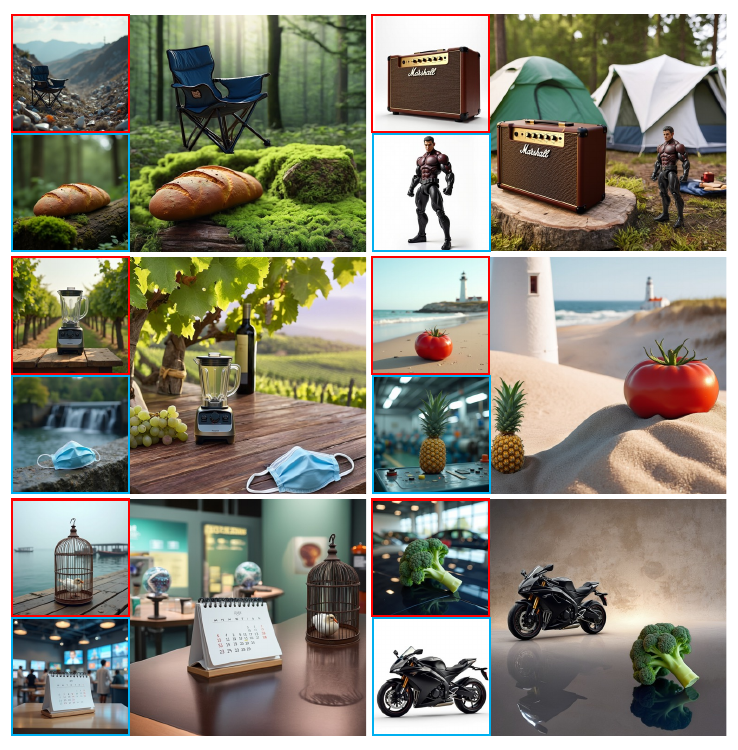}
    \caption{Showcases of PIC-98K Dataset.}
    \label{fig:samples_PIC98k}
\end{figure*}

\begin{figure*}[t]
    \centering
    \includegraphics[width=0.9\linewidth]{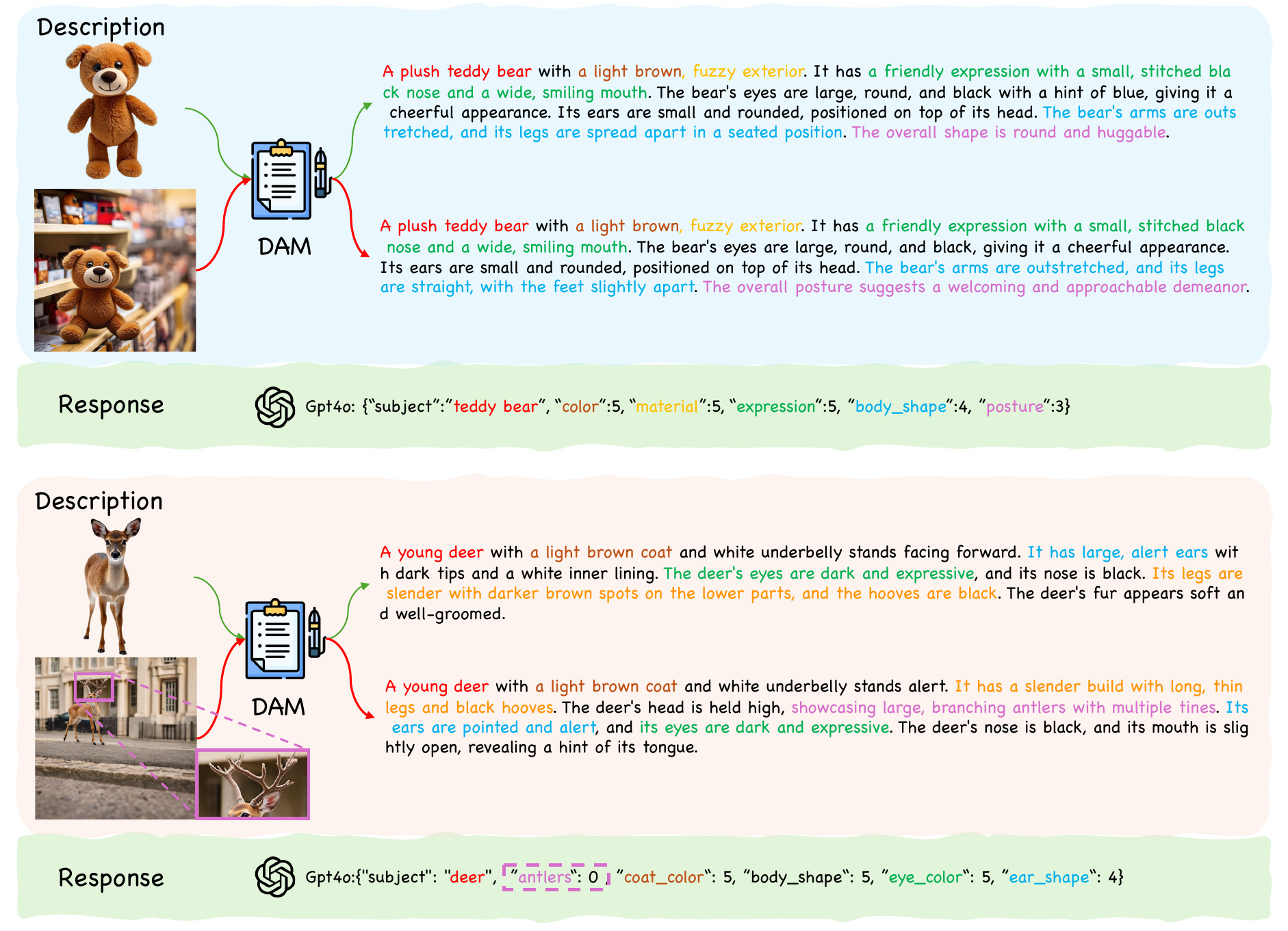}
    \caption{Examples of Multi-dimensional Perception Data Filter.}
    \label{fig:samples_gpt4o}
\end{figure*}


\end{document}